\documentclass[12pt]{class2020} 

\usepackage{tcolorbox}
\usepackage{cases}
\usepackage{bbm}
\usepackage{bbold}
\usepackage{caption, subcaption}

\title[New Potential-Based Bounds for the Geometric-Stopping Prediction with Expert Advice]{New Potential-Based Bounds for the Geometric-Stopping Version of Prediction with Expert Advice}
\usepackage{times}

\msmlauthor{%
 \Name{Vladimir A.\ Kobzar} \Email{vladimir.kobzar@nyu.edu}\\
 \addr Center for Data Science, New York University, 60 Fifth Ave., New York, New York
 \AND
 \Name{Robert V.\ Kohn} \Email{kohn@cims.nyu.edu}\and
  \Name{Zhilei Wang} \Email{zhilei@cims.nyu.edu}\\
  \addr Courant Institute of Mathematical Sciences, New York University, 251 Mercer St., New York, New York %
}

\begin{document}

\maketitle

\begin{abstract}%
This work addresses the classic machine learning problem of online prediction with expert advice. A new potential-based framework for the fixed horizon version of this problem has been recently developed using verification arguments from optimal control theory. This paper extends this framework to the random (geometric) stopping version.  To obtain explicit bounds, we construct potentials for the geometric version  from potentials used for the fixed horizon version of the problem. This construction leads to new explicit lower and upper bounds associated with specific adversary and player strategies. While there are several known lower bounds in the fixed horizon setting, our lower bounds appear to be the first such results in the geometric stopping setting with an arbitrary number of experts. Our framework also leads in some cases to improved upper bounds. For two and three experts, our bounds are optimal to leading order.    
\end{abstract}


\section{Introduction}
\label{sec:intro}  
The problem of prediction with expert advice (the \textit{expert problem}) is a classic problem in online machine learning.  We will use the following representative definition of it.  
 \begin{tcolorbox} 
  \textit{Prediction with expert advice:} 
  At each period $t \in [T]$, (a) the \textit {player} determines which of the $N$ \textit{experts} to follow by selecting a discrete probability distribution $p_t \in \Delta_N$; (b) the \textit{adversary} determines the allocation of losses to the experts by selecting a probability distribution $a_t$ over the hypercube $ [-1,1]^N$; and (c) the expert losses $q_t \in [-1,1]^N $ and the player's choice of the expert $I_t \in [N]$ are sampled from $a_t$ and $p_t$, respectively, and revealed to both parties.
 \end{tcolorbox}
 
In this setting, $a =(a_t)_{t\in [T]} $ and $p=(p_t)_{t\in [T]}$ are referred to as, respectively, the  \textit{adversary} and \textit{player strategies} or simply the adversary and player. The player strategy may be known to the adversary and vice versa. At time $t$, the player and adversary determine the probability distributions $p_t$ and $a_t$, each having the information about the outcomes of the previous periods: $\{q_\tau, I_{\tau} \}_{\tau<t}$. However, since the current period loss $q_t$ is not revealed before the player choses $p_t$ and the player's choice $I_t$ is not revealed before the adversary selects $a_t$, this implies that $q_t$ and $I_t$ are independent conditioned on the history. 

In the \textit {finite horizon} version of the expert problem, the number of periods $T$ is fixed, the regret is $R_T(p,a) =  \mathbb E_{ p,a} \left [ \sum_{t \in [T]}  (q_t)_{I_t}  - \min_{ i } \sum_{t\in [T]} (q_t)_i \right]$.  The focus of this work is the \textit{geometric stopping} version, where the final time $T$ is not fixed but is rather random, chosen from the geometric distribution $G$ with mean $\frac {1}{\delta}$, and the regret is $R(p,a) =  \mathbb E_{G} R_T(p,a)$. 
  
In the geometric setting, nonasymptotic minmax optimal strategies have not been determined explicitly, except for $N=2$ and $3$~\citep{gravin16}. Strategies that are optimal asymptotically (as ${\delta}$ approaches zero) were determined by methods based on partial differential equations (PDEs). \citet{drenska2019prediction} showed that, for any  $N$, the value function associated with a scaling limit of the geometric problem is the unique solution of an associated nonlinear PDE.   The last reference also gave a closed-form solution of the geometric stopping PDE for $N=3$.  \citet{bayraktar2019a} found a closed form solution of the geometric stopping PDE for $N = 4$, and using its inverse Laplace transform, \citet{bayraktar2019b} solved the finite horizon PDE for $N = 4$.

Extending the ideas of \citet{rakhlin2012} and \citet{rokhlin}, in the fixed horizon setting \citet{kobzar} derived player and adversary strategies using sub- and supersolutions of specific PDEs as potentials, and provided numerous examples (including lower as well as upper bounds). This  paper extends the PDE-based framework to the geometric stopping setting. In particular,  
\begin {enumerate} 
\item We provide easily-checked conditions for a function to be useful as a lower-bound or an upper bound potential (Theorems  \ref{thm:geom_lb} and \ref{thm:geom_ub}). 
 \item Using a method based on the Laplace transform, we construct potentials for the geometric problem from potentials used for the fixed horizon version (Theorems \ref{thm:fixed_to_geom_lb} and  \ref{thm:fixed_to_geom_ub}). 
\item We obtain the first known lower bounds in the geometric stopping setting with an arbitrary number of experts (Section \ref{sec:related_work}).  Our framework also leads in some cases to improved upper bounds. 
\end {enumerate} 

We use three classes of potentials to obtain explicit bounds. The first potential, discussed in Section \ref{sec:exp}, is associated with the classical exponential weights player $p^e$. The application of our framework improves the best known upper bound for $p^e$ by $\sqrt {2 \log N}$ (uniformly in $\delta$).

The second class of potentials, discussed in Section \ref{sec:screened_poisson}, is constructed by the Laplace transform of the solution to the linear heat equation.  The leading order behavior of the lower bound obtained using this potential and a randomized adversary is $\Omega \Big(\sqrt {\frac {\log N}{\delta}} \Big)$, which matches up to a prefactor the leading order behavior of the upper bound attained using exponential weights (we also provide explicit nonasymptotic guarantees). The upper bound obtained by this potential improves on the previously known bounds for small $N$ and $\delta$.

The last class of potentials, discussed in Section \ref{sec:max_fixed}, is based on closed-form solutions of a nonlinear PDE whose differential operator is the largest diagonal entry of the Hessian. The resulting bounds are optimal to leading order for $N=2$ and $3$ and also improve on previously known upper bounds for small $N$ and $\delta$. 
 
 \section{Notation} 
 
For a multi-index $I$, $\partial_{I}$ refers to the partial derivative and $dx_I$ refers to the differential with respect to the variable(s) in $I$, and $d \hat x_I$ refers to the differential with respect to all except the  variables in $I$.  $D^2 u$, $D^3 u$ and $D^4 u$  refer to the Hessian, 3rd derivative, and 4th derivative of $u$ with respect to $x \in \mathbb R^N$ (which are 2nd order, 3rd order and 4th order tensors respectively); the associated multilinear forms $\langle D^2 u \cdot q, q \rangle $, $D^3 u [q,q,q]$, $D^4 u [q,q,q,q]$ are $\sum_{i,j} \partial_{ij} u~ q_{i} q_{j} $, $\sum_{i,j,k} \partial_{ijk} u~ q_{i} q_{j} q_k $ and $\sum_{i,j,k,l} \partial_{ijkl} u~ q_{i} q_{j} q_k q_l $.

The vector $r_{\tau} = q_{I_{\tau}, \tau} \mathbb 1  -  q_{\tau}$ denotes the player's losses realized in round $\tau$ relative to those of each expert (\textit {instantaneous regret}) and the vector $x_t = \sum_{\tau <t } r_{\tau}$ denotes the player's cumulative losses realized before the outcome of round $t$ relative to those of each expert (cumulative regret or simply the \textit {regret}).

If $u$ is a function, $\Delta u = \sum_i \frac{\partial^2 u}{\partial x_i^2}$ is its Laplacian; however, the standalone symbol $\Delta_N$ refers to the set of probability distributions on $\{1, ..., N\}$.
$[T]$ denotes the set  $\{1, ..., T\}$ if $T \geq 1$ or $\{T, ..., -1\}$ if $T \leq -1$.  $\mathbb 1$ is a vector in $\mathbb R^N$ with all components equal to 1,  but $\mathbb 1_S$ refers to the indicator function of the set $S$.  

A \textit{classical solution} of a partial differential equation (PDE) on a specified region is a solution such that all derivatives appearing in the statement of the PDE exist and are continuous on the specified region.

Prediction with expert advice is a repeated two-person game. In the fixed horizon setting, it is convenient to denote the time $t$ by nonpositive numbers such that the starting time is $T \leq -1$ and the final time is zero.  The ``spatial variables'' and ``spatial derivatives'' of a function $u(x,t)$ are $x \in \mathbb R^N $ and the derivatives of $u$ with respect to $x$. 

If $u$ is a function of space and time, subscripts $x$ or $t$ denote partial derivatives (so $u_x$ and $u_t$ are first derivatives and $u_{xx}$, $u_{xt}$ and $u_{tt}$ are second derivatives). In other settings, the subscript $t$ is an index; in particular, the expert losses and player's choice at time $t$ are $q_t$ and $I_t$. When no confusion will result, we sometimes omit the index $t$, writing for example $q$ rather than $q_t$; in such a setting, $q_i$ refers to the $i$th component of $q_t$. 
 
\section {Lower Bounds} 
\label{sec:lb_geom}

Our lower bounds are associated with specific adversary strategies. We shall only consider strategies that are \textit{Markovian}, in the sense that a strategy can depend only on the cumulative regret $x$.  Such an adversary $a$ is associated with its value function $v_a(x)$, which represents the expected final time regret attained by this adversary when the prediction process starts at a cumulative regret $x$ and the player behaves optimally.  This function is characterized by the following relationship:
\begin{align}
v_a(x) =  \delta \max_i x_i +(1- \delta) \min_p \mathbb E_{a,p} v_a (x+r)
\label{eq:lb_functional_equation}
\end{align}
As this relationship determines an optimal player, that strategy is also Markovian, i.e., depends only on the cumulative regret $x$.  The function $v_a$ has an equivalent characterization:
\begin {align}
\label{eq: geom_adversary_value}
 v_a(x) &=  \min_p \mathbb E_{G, p,a} \Big[\max_i \Big (x_i+ \sum_{\tau \in [T]} r_{i,\tau} \Big) \Big]
  \end{align}
  where $T$ is sampled from the geometric distribution $G$. 

 In the context of lower bounds, we only consider those adversary strategies that assign the same expectation to each component of $q$:  $\mathbb E_{a} q_i = \mathbb E_{a} q_j$ for all $i \neq j$ (\textit{balanced adversaries}).  We associate an adversary $a$ with a potential $\hat u$ as described below.\footnote{We denote  potentials for the geometric problem with the superscript {$\hat~$} (\textit{hat}) to distinguish them from those for the fixed horizon problem.}
 \begin{tcolorbox} 
 \textit{Geometric  horizon: A lower-bound potential $\hat u$}  is a function  $\hat u: \mathbb R^N \rightarrow \mathbb  R$,  such that, for every $x \in \mathbb R^N$, there exists a balanced adversary $a$ on $ [-1, 1]^N$ ensuring that $\hat u$ is a classical solution of 
\begin{subnumcases} {\label{eq:pde_geom_lb}}
 \hat u(x)  \leq \max_i x_i + \frac {1- \delta}{ 2\delta} \mathbb E_a \langle D^2 \hat u(x) \cdot q, q \rangle  \label{eq:pde_geom_lb_ineq} \\
 \hat u(x+ c \mathbb 1)=\hat u(x)+c 
\end{subnumcases}
An \textit {adversary} associated with $\hat u$ is a balanced strategy $a$ such that \eqref{eq:pde_geom_lb_ineq} is satisfied for all $x$. 
 \end{tcolorbox}
 
 The following result shows that a lower bound potential $\hat{u}$ determines a lower bound on $v_a$. This bounds the regret below since $ v_a(0) = \min_p R(a,p)$. The bound includes an ``error" term that comes from estimating the third-order terms in the Taylor expansion of $\hat{u}$ (which is relevant because to compare $\hat{u}$ with $v_a$ we must estimate $\mathbb \min_p \mathbb E_{p,a}[\hat{u}(x+r)]$ in terms of $\hat{u}(x)$). The proof, given in Appendix  \ref{app:thm_geom_lb}, uses the characterization \eqref{eq: geom_adversary_value} of $v_a$ (as well as a characterization similar to \eqref{eq:lb_functional_equation} of a new game whose value converges to $v_a$). 
 
\begin{theorem} [Geometric horizon lower bound]
 \label{thm:geom_lb} Let $\hat u(x)$ be a lower bound potential and let $v_a$ be the value function of the associated adversary $a$. If  $\hat u(x) - \max_i x_i $ is bounded above uniformly,  then $\hat u(x)- E \leq v_a(x)$ where $E$ is an error estimate of the Taylor approximation of $\hat u$. If $D^2\hat u$ is Lipschitz continuous, then any constant $E$, satisfying $\frac {1-\delta}{6\delta} \text{ess sup}_{y \in [x,x-q]} D^3\hat u (y)[ q,q,q]   \leq  E$ uniformly for  all $q$ in the support of $a$ and all $x $, may be used. 
\end{theorem}

 Similarly to Proposition 2 in \citet{kobzar}, if the adversary assigns the same probability to $q$ and $-q$ for all $q$ in its support (a \textit{symmetric strategy}) and the potential is smooth enough,   the following alternative estimate of the error holds, as verified in Appendix \ref{app:geom_lb_lipschitz}.

\begin{proposition} [Geometric lower bound - symmetric adversary and smooth potential]
 \label{rem:geom_lb_lipschitz} If $D^3 \hat u$ exists and is Lipschitz continuous, then the conclusion of Theorem  \ref{thm:geom_lb} holds for any constant $E$ satisfying $ -  \frac {1-\delta}{24\delta}\text{ess inf}_{y \in [x,x-q]} ~D^{4}\hat u(y) [q,q,q,q]   \leq  E$ uniformly for all $q$ in the support of $a$ and all $x$. 
\end{proposition}

Next, we show how a ``stationary" fixed horizon potential can be used to determine a geometric stopping potential $\hat u$. In this setting, we denote the time $t$ by nonpositive numbers.    
\begin{tcolorbox} 
  \textit{Fixed horizon: A lower-bound potential} $u$ is a function $u: \mathbb R^N \times \mathbb R_{\leq 0}  \rightarrow \mathbb  R$,  such that, for every $x \in \mathbb R^N$, there exists some balanced probability distribution $a$ on $ [-1, 1]^N$ that depends on $x$ only (i.e., it is \textit{stationary}) ensuring that $u$ is a classical solution of 
\begin{subnumcases} { \label{eq:pde_fixed_lb}}
 u_t +  \frac {1- \delta}{ 2\delta}   \mathbb E_{a} \langle D^2 u \cdot q, q \rangle \geq 0 \\
 u(x,0) = \max_i x_i \\
  u(x+ c \mathbb 1, t)=u(x, t)+c 
\end{subnumcases}
 \end{tcolorbox}
 
Consider a potential $\hat u$ given by 
\begin{align}
\label{eq:lb_transform}
\hat u(x) = e^{\delta} \int_{- \infty }^{-\delta} e^{t} u(x,t) dt - C
\end{align}
 where $C$ is a constant that such  $u(x, -\delta) -  \max_i x_i  \leq C $ for all $x$.  Integrating by parts,
\begin{align*}
\hat u (x) &= e^{\delta} \int_{- \infty }^{-\delta} \partial_t e^t u(x,t) dt - C = u(x,-\delta)- C- e^{\delta}  \int_{- \infty }^{-\delta} e^t u_t(x,t) dt\\
 & \leq u(x,-\delta)- C+ e^{\delta}  \int_{- \infty }^{-\delta} e^t \left (\frac {1- \delta}{ 2\delta}   \mathbb E_{a} \langle D^2 u \cdot q, q \rangle \right) dt  \leq  \max_i x_i+\frac {1- \delta}{ 2\delta}  \mathbb E_{a}  \langle D^2 \hat u \cdot q, q \rangle
\end{align*}
where the last inequality holds because $a$ is stationary.   Also the linearity of $u$ in the direction of $\mathbb 1$  implies the same result with respect to $\hat u$. Therefore,  $\hat u$ associated with the adversary strategy $a$ satisfies \eqref{eq:pde_geom_lb}. 

If $\mathbb E_{a} \langle D^2 u \cdot q, q \rangle$ is bounded above uniformly in $x \in \mathbb R^N$, $t\leq -\delta$,  then $ \hat u(x) - \max_i x $  is uniformly bounded above.  Finally, if for any $q$ in the support of $a$, $\text{ess sup}_{y \in [x,x-q]} D^3 u (y, t)[ q,q,q]   \leq  C_3(t)$, then
\begin{align*}
 \text {ess~sup}_{y \in [x,x-q]}  D^3\hat u (y)[ q,q,q]   \leq e^{\delta} \int_{- \infty }^{-\delta}  e^{t}C_3(t) dt  
\end{align*}
Thus, this calculation obtains a geometric lower bound potential associated with the same adversary as the fixed horizon lower bound. 
\begin{theorem} [Fixed horizon to geometric lower bound]
 \label{thm:fixed_to_geom_lb}  Let $u$ be a fixed horizon lower bound potential satisfying \eqref{eq:pde_fixed_lb} and associated with a stationary adversary $a$ and let $v_a$ be the value function of this adversary in the geometric stopping problem.  If $u(x, -\delta) -  \max_i x_i  \leq C$ uniformly for all $x$  and  $\mathbb E_{a} \langle D^2 u(x, t) \cdot q, q \rangle$  is bounded above uniformly for all $x$ and $t \leq -\delta$, then $\hat u(x)- E \leq v_a(x)$ where $\hat {u}$ is given by \eqref{eq:lb_transform} and $E$ is an error estimate of the Taylor approximation of $\hat u$. If $D^2 u(\cdot, t)$ is Lipschitz continuous and  $\text{ess sup}_{y \in [x,x-q]} D^3 u (y, t)[ q,q,q]  \leq  C_3(t)$ uniformly in $x$ and all $q$ in the support of $a$, then $E =\frac {1-\delta}{6\delta} e^{\delta} \int_{- \infty }^{-\delta}  e^{t}C_3(t) dt$ may be used. 
\end{theorem}

 Similarly to Proposition  \ref{rem:geom_lb_lipschitz}, if the fixed horizon potential is smooth enough and the associated strategy is symmetric, we can obtain an alternative error estimate. Specifically, if $D^3 u (\cdot, t)$ exists and is Lipschitz continuous, and for all $t<0$, $- \text {ess~inf}_{y \in [x,x-q]}   D^4 u (y, t)[q, q,q,q]   \leq  C_4(t)$, then
\begin{align*}
- \text {ess~inf}_{y \in [x,x-q]}  D^4 \hat u (y)[ q,q,q,q] & \leq e^{\delta} \int_{- \infty }^{-\delta} e^{t}C_4(t) dt 
\end{align*} 
leading to the following proposition. 
\begin{proposition} [Fixed horizon to geometric l.b. - symmetric adversary and smooth potential]
 \label{rem:fixed_to_geom_lb_lipschitz} 
In the context of Theorem \ref{thm:fixed_to_geom_lb},  if  $D^3 u (\cdot, t)$ exists and is Lipschitz continuous, and for all $t<0$,\\ $- \text {ess~inf}_{y \in [x,x-q]}   D^4 u (y, t)[q, q,q,q]   \leq  C_4(t)$ uniformly for all $q$ in the support of $a$ and all $x$, then $E = \frac {1-\delta}{24\delta} e^{\delta} \int_{- \infty }^{-\delta}  e^{t}C_4(t) dt$ may be used. 
\end{proposition}

\section {Upper Bounds} 
\label{sec:ub_geom}

Our upper bounds are associated with well-chosen potentials for the player. We shall also only consider potentials that are Markovian, i.e., that depend only on the cumulative regret $x$.  The associated player $p$ is given by the gradient of such potential, and the value function $v_p(x)$ of $p$ represents the expected final time regret attained by this player when the prediction process starts at a cumulative regret $x$ and the adversary behaves optimally. This function is characterized by the following relationship 
\begin{align}
v_p(x) =  \delta \max_{i} x_i + (1-\delta) \max_a  \mathbb E_{a, p} ~v_p(x+r)
\label{eq:ub_functional_equation} 
\end{align}
 The function $v_p$ also has an equivalent characterization:
\begin {align}
 v_p(x) =  \max_a \mathbb E_{G, p,a} \max_i \Big [\Big (x_i+ \sum_{\tau \in [T]} r_{\tau,i} \Big) \Big ]
  \label{eq: geom_player_value}
\end{align}
To bound $v_p$ from above, we introduce an upper bound potential $\hat w$ and the corresponding player $p$.
\begin{tcolorbox} 
  \textit{Geometric stopping: an upper bound potential} $\hat w$ is a function $\hat w: \mathbb R^N   \rightarrow \mathbb  R$, which is nondecreasing as a function of each $x_i$, and which, for all $x \in \mathbb R^N$, and $c \in \mathbb R$, is a classical solution of 
\begin{subnumcases} {\label{eq:pde_geom_ub}}
\hat w(x)  \geq \max_i x_i + \frac {1- \delta}{ 2\delta}  \max_{q \in [- 1,1]^N}\langle D^2 \hat w(x) \cdot q, q \rangle\\
\hat w (x+c\mathbb 1) = \hat w (x)+c 
\end{subnumcases}
The associated \textit{player strategy} $p$ associated with $\hat w$ is given by $p =\nabla \hat w(x)$.   
 \end{tcolorbox}
\noindent  Since $\hat w$ is nondecreasing as a function of each  $x_i$ and $\sum_i \partial_i \hat w(x) =1$ by linearity of $\hat w(x)$ in the direction of $\mathbb 1$, $p \in \Delta_N$. 

The following result shows that an upper bound potential $\hat{w}$ determines an upper bound on $v_p$. This bounds the regret above since $ v_p(0) = \max_a R(a,p)$. This bound also includes an ``error" term that comes from estimating the third-order terms in the Taylor expansion of $\hat{w}$ (which is relevant because to compare $\hat{w}$ with $v_p$ we must estimate $\max_a \mathbb E_{p,a}[\hat{w}(x+r)]$ in terms of $\hat{w}(x)$). The proof, given in Appendix  \ref{app:thm_geom_ub}, uses the characterization \eqref{eq: geom_player_value} of $v_p$ (as well as a characterization similar to \eqref{eq:ub_functional_equation} of a new game whose value converges to $v_p$). 

\begin{theorem} [Geometric horizon upper bound]
\label{thm:geom_ub} Let $\hat w(x)$ be an upper bound potential and let $v_p$ be the value function of the associated player. If  $\hat w(x) - \max_i x_i $ is uniformly bounded below,  then $ v_p(x) \leq \hat w(x) +E$ where $E$ is an error estimate of the Taylor approximation of $\hat w$. If $D^2\hat w$ is Lipschitz continuous, then any constant $E$, satisfying $-\frac {1-\delta}{6\delta} \text{ess inf}_{y \in [x,x-q]} D^3\hat w (y)[ q,q,q]  \leq  E$ uniformly for  all $q \in [-1,1]^N$  and all $x$, may be used. 
 \end{theorem}

We can also construct a geometric stopping upper bound potential from a fixed horizon one. 

\begin{tcolorbox} 
  \textit{Fixed horizon: an upper-bound potential} $w$ is a function $w: \mathbb R^N \times \mathbb R_{\leq 0}  \rightarrow \mathbb  R$, which is nondecreasing as a function of each $x_i$, and which is, for all $x \in \mathbb R^N$, $t <0$, and $c \in \mathbb R$, a classical solution of 
\begin{subnumcases} { \label{eq:pde_fixed_ub}}
 w_t + \frac {1- \delta}{ 2\delta} \max_{q \in [- 1,1]^N} \langle D^2 w \cdot q, q \rangle \leq 0 \label{eq:pde_ineq_fixed_ub} \\
 w(x,0) \geq \max_i x_i \label{eq:pde_final_fixed_ub} \\
 w (x+c\mathbb 1,t) = w (x,t)+c  
\end{subnumcases}
 \end{tcolorbox}
 
Consider a potential $\hat w$ given by 
\begin{align}
\label{eq:ub_transform}
\hat w(x) = e^{\delta} \int_{- \infty }^{-\delta} e^{t} w(x,t) dt + C
\end{align}
 where $C$ is a constant that such  $\max_i x_i - w(x, -\delta)   \leq  C $ for all $x$.  Integrating by parts,
\begin{align}
\hat w (x) &= e^{\delta} \int_{- \infty }^{-\delta} \partial_t e^t w(x,t) dt + C = w(x,-\delta)+ C- e^{\delta}  \int_{- \infty }^{-\delta} e^t w_t(x,t) dt \nonumber\\
 & \geq w(x,-\delta)+ C+ e^{\delta}  \int_{- \infty }^{-\delta} e^t \left (\frac {1- \delta}{ 2\delta}   \max_{q \in [-1,1]^N} \langle D^2 w \cdot q, q \rangle \right) dt \\
 & \geq \max_i x_i+\frac {1- \delta}{ 2\delta}  \max_{q \in [-1,1]^N } \langle D^2 \hat w \cdot q, q \rangle
 \label{eq:ub_potential_laplace}
\end{align}
 Also the linearity of $w$ in the direction of $\mathbb 1$  implies the same result with respect to $\hat w$. Also, if $w$ is nondecreasing with respect to each $x_i$, then so is $\hat w$.   Therefore,  $\hat w$ satisfies \eqref{eq:pde_geom_ub}. 
 
 If $ \langle D^2 w \cdot q, q \rangle$ is bounded below uniformly in  $q \in [- 1,1]^N$, $x \in \mathbb R^N$ and $t\leq -\delta$, then $ \hat w(x) - \max_i x $  is uniformly bounded below.  
Finally, if for any $q$, $-\text{ess inf}_{y \in [x,x-q]} D^3 w (y, t)[ q,q,q]   \leq  C_3(t)$, then
\begin{align*}
 -\text {ess~inf}_{y \in [x,x-q]}  D^3\hat w (y)[ q,q,q] \leq e^{\delta} \int_{- \infty }^{-\delta}  e^{t}C_3(t) dt  
\end{align*}
Thus, this calculation obtains a geometric upper bound potential from the fixed horizon one. 

\begin{theorem} [Fixed horizon to geometric upper bound]
 \label{thm:fixed_to_geom_ub} Let $w$ be a fixed horizon upper bound potential satisfying \eqref{eq:pde_geom_ub}.  If $ \max_i x_i- w(x, -\delta)    \leq C$ uniformly for all $x$  and  $\max_{q \in [-1,1]^N}\langle D^2 w \cdot q, q \rangle$  is bounded below uniformly for all $t \leq -\delta$ and all $x$, then $v_p(x) \leq \hat w(x)+ E $ where $\hat {w}$ is given by \eqref{eq:ub_transform}, $v_p(x)$ is the value function associated with the player $p = \nabla \hat w(x)$ and $E$ is an error estimate of the Taylor approximation of $\hat w$. If $D^2 u$ is Lipschitz continuous and  $-\text{ess inf}_{y \in [x,x-q]} D^3 u (y, t)[ q,q,q]  \leq  C_3(t)$ uniformly for all $q \in [-1,1]^N$ and all $x$, then $E =\frac {1-\delta}{6\delta} e^{\delta} \int_{- \infty }^{-\delta}  e^{t}C_3(t) dt$ may be used.  
\end{theorem}

The analysis becomes simpler if a fixed horizon upper bound potential has the form
\begin{align}
w(x,t) =  \Phi(x)+kt
\label{eq:stationary_players}
\end{align} for a constant $k$. As shown in Appendix \ref{app:fixed_to_geom_ub_const_w_t}, this allows us to use Taylor's expansion of the resulting geometric upper bound potential $\hat w$ with the mean value form of the second-order  remainder, which eliminates the discretization error $E$. (Also, in such case, we do not need to ensure convergence of the integral for the error term $E$, and therefore we can take the upper limit of integration to be $0$ rather than $\delta$.)  

\begin{proposition} [Fixed to geometric upper bound - certain potential]
 \label{rem:fixed_to_geom_ub_const_w_t}  In the context of Theorem \ref{thm:fixed_to_geom_ub}, if $w$ has the form given by \eqref {eq:stationary_players}, then  $v_p(x) \leq \hat w(x)$ where $\hat w(x) =  \int_{- \infty }^{0} e^{t} w(x,t) dt + C$ and $C$ is a constant that such  $\max_i x_i - w(x, 0)   \leq  C $ for all $x$.
\end{proposition}
In the next section, we will use this Proposition to derive an upper bound for the exponentially weighted average player.

\section{Exponential weights}
\label{sec:exp}

The fixed horizon upper bound potential $w^e$ given by $w^e(x,t) =\Phi(x) -  \frac {1- \delta}{2 \delta} \eta t$ where $\Phi(x) = \frac{1}{\eta} \log(\sum_{k=1}^N e^{\eta x_k})$ is associated with the exponential weights player $p^e$. By a standard result, $ \langle D^2 \Phi \cdot q,q\rangle   \leq \eta$ for all $q \in [-1,1]^N$ and all $x$.\footnote{See, e.g., Appendix E in \citet{kobzar}} Also note that $\Phi (x+c\mathbb 1) =  \Phi (x)+c$. Accordingly, $w^e$ satisfies \eqref{eq:pde_fixed_ub}. Let the corresponding geometric potential $\hat w^e$ be given by 
\[
\hat w^e(x) = \int_{- \infty }^{0} e^{t} w^e(x,t) dt = \Phi(x) + \frac {1- \delta}{2 \delta} \eta
\]  
Note that $ \max_i x_i- w^e(x,0)  \leq 0 $. Also since $\Phi$ is convex,  $0 \leq \langle D^2 \Phi \cdot q,q\rangle$.  Therefore, by Proposition \ref{rem:fixed_to_geom_ub_const_w_t}, we obtain
\begin {example} [Exponential weights] 
\label{ex:exp_ub} The exponential weights player $p^e$ attains the following upper bound $v_{p^e}(x) \leq \hat w^e(x)  $ where  $v_{p^e}$ is the value function for this player and  $\hat w^e(0)   =\frac {\log N}{\eta} + \frac {1- \delta}{2 \delta} \eta$.  Taking $\eta = \sqrt {\frac {2 \delta \log N}{1 - \delta}}$ leads to the  regret bound $\max_a R(a,p^e) \leq \sqrt {\frac {2(1- \delta) \log N}{ \delta}}$. 
\end{example}

\section {Heat Potentials} 
\label{sec:screened_poisson}
In this section, we consider the function $\varphi$ given by 
\begin{align}
\label{eq:heat_potential}
\varphi(x,t) &= \alpha \int  e^{-\frac  {\|y \|^2} {2\sigma^2} }\max_k (x_k-y_k) dy
\end{align}
where $\alpha = (2 \pi \sigma^2)^{-\frac{N}{2}}$ and $\sigma^2 = -2 \kappa t$ and $t<0$. This function is the classical solution, on $\mathbb R^N \times \mathbb R_{<0}$, of the following linear heat equation
\begin{align*}
\begin{cases}
u_t +\kappa \Delta u =0  \\
u(x,0) = \max_i x
\end{cases}
\end {align*}

We also consider a function $\hat \varphi$ given by 
\begin{align*}
\hat \varphi(x) = e^{\delta} \int_{- \infty }^{-\delta} e^{t} \varphi(x,t) dt 
\end{align*}
Since $\frac{d}{dt} (e^t \sqrt{-t} - \frac{\pi }{2} \text{erf} (\sqrt {-t}) = e^t \sqrt {-t}$,  we obtain 
\begin{align}
\label{eq: sqrt_int}
 \int_{- \infty }^{-\delta} e^t \sqrt {- t} dt = e^{-\delta} \sqrt {\delta} + \frac {\sqrt \pi}{2} (1- \text{erf} (\sqrt {\delta}))
\end{align}
 Therefore,
\begin {align}
\hat \varphi(0) &= \sqrt {2\kappa } \mathbb E_Y \max Y_i \Big (e^{\delta} \int_{- \infty }^{-\delta} e^t \sqrt {- t} dt \Big)  = \sqrt {2\kappa } \mathbb E_Y \max Y_i \Big ( \sqrt {\delta} + e^{\delta}\frac {\sqrt \pi}{2} \text{erfc} (\sqrt {\delta})) \Big ) 
\label{eq:varphi_at_zero}
\end{align}


By Section 5 of \citet{kobzar}, the function $\varphi$ defined by \eqref{eq:heat_potential} with the diffusion factor 
\begin {align*}
\kappa_s  = \frac {1- \delta}{\delta} ~ \text{if}~ N =2\text {,}~
 \frac {1- \delta}{\delta} \Big(\frac{1}{2} + \frac {1}{2N}\Big) ~\text{if}~ N ~\text{ is odd,} ~ \text {or}~
 \frac {1- \delta}{\delta} \Big (\frac{1}{2} + \frac {1}{2N-2} \Big) ~\text{otherwise}
\end{align*}
satisfies \eqref {eq:pde_fixed_lb} when the adversary uses the following strategy.
 \begin{tcolorbox}
   \textit{Heat adversary} $a^h$ samples $q$ from a uniform distribution on the set S:
\[
S=   
\Big \{q\in\{\pm 1\}^N\mid \sum_{i=1}^Nq_i=\pm 1 \Big\} ~\text{for}~ N~\text{ odd} ~~~\text{and}~~~ \Big \{q\in\{\pm 1 \}^N\mid \sum_{i=1}^Nq_i=0 \Big \} ~\text{for}~ N~\text{even}.
\]
 \end{tcolorbox}

Note that the strategy $a^h$  is symmetric and stationary. Next, as confirmed in Appendix \ref{heat_final_period},  $|\varphi(x, -\delta) -  \max_i x_i | \leq C^\varphi$ for all $x \in \mathbb R^N$ where $C^\varphi =  \sqrt  {2 \kappa \delta } \mathbb E \max_i Y_i$, and $Y$ is an $N$-dimensional Gaussian random vector with mean zero and identity covariance matrix.  Also by Appendices I and F.1 in \citet{kobzar}, for  $\varphi$ with $\kappa = \kappa_s$, 
\begin {align*}
\max_q \langle D^2 \varphi \cdot q, q \rangle  \leq 2 \Delta \varphi &= -\sum_i \frac{2\alpha}{\sigma^2}\int e^{-\frac{\|y\|^2}{2
\sigma^2}}y_i\mathbb{1}_{x_i-y_i>\max_{j\neq i}x_j-y_j}dy =  \frac{2}{\sigma} \mathbb E_Y \max_i |Y_i|\\
& = \frac{2}{\sqrt {-2\kappa_s t }} \mathbb E_Y \max_i |Y_i| \leq  \frac{2}{\sqrt {2(1- \delta) }}\mathbb E_Y \max_i |Y_i|
\end{align*}
for $t \leq -\delta$. In particular, $ \max_q \langle D^2  \varphi \cdot q, q \rangle$  is uniformly bounded above.  Therefore,   $\varphi $ with the diffusion factor $\kappa = \kappa_s$ associated with the adversary $a^h$ is a fixed horizon lower bound potential satisfying the conditions of Theorem \ref{thm:fixed_to_geom_lb}. The resulting geometric lower bound potential is given by  $\hat u^h =  \hat \varphi -C^\varphi $ where $\hat \varphi$ has the same diffusion factor. 

Let $E_{\hat u^h} $ denote the error term within the meaning of Theorem \ref{thm:fixed_to_geom_lb} for $\hat u^h$  associated with  the adversary $a^h$.  By Appendix F.2 in \citet{kobzar}, $|D^3 \varphi (y, t)[ q,q,q] | \leq \frac{K_3}{|t|}$ for all $q$ where $K_3 =  O(\kappa^{-1}\sqrt N)$.  Using a standard bound of the exponential integral, this error term can be computed as follows:
\begin{align}
 \frac {1-\delta}{6\delta} e^{\delta} \int_{- \infty }^{-\delta}  e^{t}\frac{K_3}{|t|} dt  \leq  \frac {1-\delta}{6\delta}  K_3 \Big(1+ \log {\frac {1}{\delta}}\Big) =E_{\hat u^h} 
 \label{eq:third_deriv_screened}
\end{align}
Taking $\kappa = \kappa_s$,  $E_{\hat u^h}=  O\left (\sqrt{N} \left(1+ \log \frac {1}{\delta} \right) \right)$. 

Since $a^h$ is symmetric and $\varphi$ is smooth, we can also use  Proposition \ref{rem:fixed_to_geom_lb_lipschitz} to compute $E_{\hat u^h}$.  By Appendix F.2 in \citet{kobzar},  $|D^4 \varphi (y, t)[q, q,q,q] | \leq  \frac{K_4}{|t|^{\frac{3}{2}}}$ for an adversary supported on $\{\pm 1\}^N$, such as $a^h$, where $K_4 = O(\kappa^{-\frac 3 2}N\sqrt N)$.  Therefore, since $\frac{d}{dt} \Big[2 \sqrt {\pi} \text{erf}(\sqrt {-t})+ \frac{2 e^t}{\sqrt {-t}} \Big] =  \frac{e^{t}}{(-t)^{\frac{3}{2}}}$, we have 
\begin{align*}
  \frac {1-\delta}{24\delta} e^{\delta}K_4 \int_{- \infty }^{-\delta}  \frac{e^{t}}{|t|^{\frac{3}{2}}} dt
 = \frac {1-\delta}{24\delta} K_4\left(  \frac {2}{\sqrt{\delta}}-2e^{\delta}  \sqrt {\pi} \text{erfc} (\sqrt {\delta})   \right) = E_{\hat u^h}
\end{align*}
where $ E_{\hat u^h} =  O\left ( N\sqrt N \right)$.  Combining these estimates, $ E_{\hat u^h} =  O\left ( N\sqrt{N} \wedge \sqrt{N} \left  (1+ \log \frac {1}{\delta} \right) \right)$.

For the upper bound, Section 5 in \citet{kobzar} provides that when $\varphi$ is defined using  $\kappa = \frac{1- \delta}{\delta} $, it satisfies \eqref{eq:pde_fixed_ub}. As noted previously, $|\varphi(x, -\delta) -  \max_i x_i | \leq C^\varphi$ for all $x \in \mathbb R^N$.  Also since $\max $ is convex, $\varphi$ is convex as a function of $x$. This implies that  $\langle D^2 \varphi \cdot q, q \rangle $ is bounded below by zero.  Therefore,   $\varphi $ with the above mentioned diffusion factor is a fixed horizon upper bound potential satisfying the conditions of Theorem \ref{thm:fixed_to_geom_ub}. The resulting geometric upper bound potential $\hat w^h=\hat \varphi+C^\varphi$ with $\kappa = \frac{1- \delta}{\delta} $, and the corresponding strategy is
 \begin{tcolorbox}
 The \textit{heat player} $p^h$ selects $p^h = \nabla \hat w^h(x)$ in each period. 
 \end{tcolorbox}
 
Let $E_{\hat w^h}$ denote the error term within the meaning of Theorem \ref{thm:fixed_to_geom_ub} for $\hat w^h$.  By a calculation similar to \eqref{eq:third_deriv_screened}, $E_{\hat w^h} =  O\left (\sqrt{N} \left(1+ \log \frac {1}{\delta} \right) \right)$. The following example summarizes these results.

\begin {example} [Heat-based strategies] 
\label{ex:poisson_bounds} The value function $v_{a^h}$ of  $a^h$ attains the lower bound $\hat u^h(x) - E_{\hat u^h} \leq v_{a^h}(x)$ and  the value function $v_{p^h}$ of $p^h$  attains the  upper bound $v_{p^h}(x) \leq \hat w^h(x) + E_{\hat w^h}$ where the potentials  $\hat u^h$ and $\hat w^h$, and error terms $E_{\hat u^h}$ and $E_{\hat w^h}$ are given above. By equation \eqref{eq:varphi_at_zero}, these bounds lead to the following regret bounds $\sqrt {2\kappa_s} \mathbb E_Y \max Y_i \left (e^{\delta} \frac {\sqrt {\pi}} {2}  \text{erfc} (\sqrt {\delta}) \right ) -  E_{\hat u^h} \leq   \min_p R(a^h,p)$ and  $\max_a R(a,p^h)  \leq \sqrt { \frac{2(1- \delta)}{\delta} } \mathbb E_Y \max Y_i \left (e^{\delta} \frac {\sqrt {\pi}} {2} \text{erfc} (\sqrt {\delta}) + 2 \sqrt {\delta} \right )+E_{\hat w^h}$.
\end{example}

For $N =  2$, $E_Y \max Y_i  = \frac {1}{\sqrt \pi}$, and therefore, $\lim_{\delta \rightarrow 0} \hat u^h(0) = \lim_{\delta \rightarrow 0}\hat w^h(0) =  \frac {1} {\sqrt {2 \delta}}$. Since the lower and upper bounds match to leading order, they are asymptotically optimal.  This result also matches  the leading order of the exact minimax regret for two experts determined in Theorem 4.1 of \citet{gravin16} (as rescaled for our range of losses). 

\section{Max potentials}
\label{sec:max_fixed}

In this section, let $\psi $ be given by the solution of 
\begin{align}
\label{eq:max_pde}
\begin{cases}
\psi_t + \kappa \max_{i}  \partial_i^2 \psi =0  \\
\psi (x,0) = \max_i x
\end{cases}
\end {align}
For all $x \in \mathbb R^N$, we will denote by $\{ (i) \}_{i \in [N]}$ the ranked coordinates of $\mathbb R^N$ such that $x_{(1)} \geq x_{(2)} \geq ... \geq x_{(N)}$.   By Claim 5 in \citet{kobzar}, the PDE \eqref{eq:max_pde} has an explicit classical solution on $\mathbb R^N \times \mathbb R_{<0}$ given by
 \begin{align*} 
\psi(x,t) &= \frac {1}{N} \sum_i x_{(i)} + \sqrt{-2\kappa t } \sum_{l=1}^{N-1} c_l f(z_l) 
\end{align*}
where $ z_l = \frac {1}{\sqrt {-2\kappa t}} \left ( \left (\sum_{n=1}^l{x_{(n)}} \right) - lx_{(l+1)} \right)$, $f(z) = \sqrt{\frac {2}{ \pi}} e^{-\frac{z^2}{2}} +z \text{erf} \left ( \frac{z}{\sqrt{2}} \right)$, $\text{erf}(y)  = \frac {2}{\sqrt \pi }  \int_0^{y } e^{-s^2} ds$ and $c_l = \frac{1}{l(l+1)}$. We consider a function $\hat \psi$ given by 
\begin{align*}
\hat \psi (x) = e^{\delta} \int_{- \infty }^{-\delta} e^{t} \psi (x,t) dt 
\end{align*}
Since $f(0) =\sqrt \frac{ 2}{\pi}$ and $\sum_{l=1}^{N-1} c_l = \frac{N-1}{N}$,  we obtain $\psi(0,t) =\frac {2(N-1)}{N}  \sqrt {-\frac {\kappa t}{\pi} }$. Using \eqref{eq: sqrt_int},
 \begin {align}
\hat \psi (0) &=\frac {2(N-1)}{N}  \sqrt {\frac {\kappa }{\pi} } \left (e^{\delta} \int_{- \infty }^{-\delta} e^t \sqrt {- t} dt  \right)= \frac {(N-1)}{N}  \sqrt { {\kappa}} \left (e^{\delta}  \text{erfc} (\sqrt {\delta}) +   \frac {2 }{\sqrt{\pi}} \sqrt {\delta}\right )
\label{eq:psi_at_zero}
\end{align}


As confirmed in Appendix \ref{max_final_period}, $\psi(x, -\delta) -  \max_i x_i  \leq C^{\psi}$ for all $x \in \mathbb R^N$ if $C^{\psi} = 2\sqrt {\frac{\kappa \delta }{\pi}}  \frac {N-1}{N}$. Also the expressions for the derivatives of  $\psi$ in Appendix J of \citet{kobzar} indicate that $\max_{i}  \partial_{ii}  \psi(x,t)$ is uniformly bounded above for all $x$ and $t \leq -\delta$.

Section 6 of \citet{kobzar} confirms $\psi$ with $\kappa =  \frac{2(1- \delta)}{\delta} $ satisfies \eqref{eq:pde_fixed_lb}  for the following adversary $a^m$. 

\begin{tcolorbox} The \textit{max adversary} $a^m$ assigns probability $\frac {1} {2}$ to each of $q^m $ and $-q^m$ where the entry of $q^m$ corresponding to the largest component of $x$ is set to 1 and the remaining entries are set to $-1$.
 \end{tcolorbox} 
 
 Therefore,    $\psi $ with the diffusion factor $\kappa =   \frac{2(1- \delta)}{\delta}$ associated with $a^m$ is a fixed horizon lower bound potential satisfying the conditions of Theorem \ref{thm:fixed_to_geom_lb}. The resulting geometric lower bound potential $\hat {u}^m =\hat \psi + C^{\psi} $. (We shall call this potential and $\hat w^m$,  below the \textit{max potentials}.)

Let $E_{\hat {u}^m} $ denote the error term within the meaning of  Theorem \ref{thm:fixed_to_geom_lb} for $\hat {u}^m$. By Appendix M.1 in \citet{kobzar}, $\text {ess~sup}_{y \in [x,x \mp q^m]} \pm D^3\hat \psi (y)[ q^m,q^m,q^m]  \leq \frac{K^{l.b.}_3}{|t|}$ where $K^{l.b.}_3 =  O(\kappa^{-1} N)$.   By a calculation similar to \eqref{eq:third_deriv_screened}, $E_{\hat {u}^m} =O\left (N \left(1+ \log \frac {1}{\delta} \right) \right)$.

To determine an upper bound,  Section 6 of \citet{kobzar} shows that $\psi $ with the diffusion factor 
\begin{align}
\kappa_m= \frac {1- \delta}{\delta} \frac{N^2}{2(N-1)} ~\text{for}~ N~\text{even,} ~\text {or}~ \frac {1- \delta}{\delta}\frac{N+1}{2} ~ \text{for}~ N~\text{odd}
\label{eq:lower_bound_max_factor}
\end{align}
satisfies \eqref{eq:pde_fixed_ub}.  Also Appendix \ref{max_final_period} below confirms that  $\psi(x, -\delta) -  \max_i x_i  \geq \hat 0$, and examination of Appendix J in \citet{kobzar} reveals that $\max_{i}  \partial_{ii}  \psi$ is uniformly bounded below for all $x$ and $t \leq -\delta$.    
Therefore,  $\psi $ with the  diffusion factor $\kappa_m$ is a fixed horizon upper bound potential satisfying the conditions of Theorem \ref{thm:fixed_to_geom_ub}. The resulting geometric upper bound potential is $\hat w^m=\hat \psi$ with $\kappa = \kappa_m $ , and the corresponding strategy is \begin{tcolorbox}
 The \textit{max-based player} $p^m$ selects $p^m = \nabla \hat w^m(x)$. 
 \end{tcolorbox}
 
Let $E_{\hat w^m } $ denote the error term within the meaning of Theorem \ref{thm:fixed_to_geom_ub} for $\hat w^m$.  By Appendix M.2 in  \citet{kobzar},  $| D^3 \psi (y)[ q,q,q]| \leq   \frac{K^{u.b.}_3}{|t|}$ for all $q \in [-1,1]^N$ where $K^{u.b.}_3 =  O(\kappa^{-1} N^2)$.   By a calculation similar to \eqref{eq:third_deriv_screened}, $E_{\hat w^m} =O\left (N \left(1+ \log \frac {1}{\delta} \right) \right)$. The next example summarizes these results.

 \begin {example} [Max-based  strategies]
\label{ex:max_bound}
The value function $v_{a^m}$ of $a^m$ satisfies the lower bound $ \hat u^m(x) - E_{\hat u^m }  \leq v_{a^m}(x)$, and the value function $v_{p^m}$ of $p^m$ satisfies the upper bound $  v_{p^m}(x)\leq \hat w^m(x)+ E_{\hat w^m } $ where the potentials $\hat u^m$ and $\hat w^m$, and the error terms $E_{\hat u^m }$ and $E_{\hat w^m }$ are as defined above.  By equation \eqref{eq:psi_at_zero}, these bounds lead to the following regret bounds $\frac {(N-1)}{N}  \sqrt {  \frac{2(1- \delta)}{\delta } } \left (e^{\delta}   \text{erfc} (\sqrt {\delta}) \right )   -   E_{\hat u^m  } \leq  \min_p R(a^m,p)$ and $
\max_a R(a,p^m)  \leq  \frac {(N-1)}{N}  \sqrt { {\kappa_m}} \left (e^{\delta}  \text{erfc} (\sqrt {\delta}) +   \frac {4 }{\sqrt{\pi}} \sqrt {\delta}\right ) + E_{\hat w^m} $. 
\end{example}
 
 For $N =  2$,  $\lim_{\delta \rightarrow 0} \hat u^m(0) = \lim_{\delta \rightarrow 0}\hat w^m(0) =  \sqrt {\frac {1-\delta} {2 \delta}}$, and  for $N =  3$,  $\lim_{\delta \rightarrow 0} \hat u^m(0) = \lim_{\delta \rightarrow 0}\hat w^m(0) =  \frac {4}{3} \sqrt {\frac {1-\delta} {2 \delta}}$. Since the lower and upper bounds match to leading order, they are asymptotically optimal.   They also match  the leading order of the exact minimax regret for two and three experts determined in Theorems 4.1 and 4.2 of \citet{gravin16} (as rescaled for our range of losses). 

\section {Related Work}
\subsection {PDE Characterizing Asymptotically Optimal Value}

The max potential was known from \citet{drenska2019prediction} to solve the PDE associated with the scaling limit of the optimal value function for $N=3$ in the geometric case. The upper and lower bounds we obtained match at the leading order for $N =2$ and $3$, and therefore solve that PDE.   

For general $N$, while our framework does not rely on the PDE characterizing asymptotically optimal value, it has the following relationship to that PDE.   Let $v$ be the minimax optimal value of the game, and let $v^{\delta}(x)  =\sqrt {\delta} v \left (\frac{x}{\sqrt {\delta}}\right )$ be a scaled version of this value function. The viscosity solution of the PDE (3.8) in \citet{drenska2019prediction}, which we denote by $v^*$, represents a scaling limit of $v^{\delta}$:  $v^*(x)  = \lim_{\delta \rightarrow 0}v^{\delta} (x)$. This PDE, as adjusted to the losses in the present paper, is given by\footnote{Separately from the scaling by $\delta$, discussed in the text accompanying this footnote,  the losses in the original unscaled game in \citet{drenska2019prediction}  are selected from $\{0,1\}^N$, rather than  $[-1,1]^N$ in the present paper. This distinction, however, is not consequential.}
\begin{align}
\label{eq:drenska_pde_geom}
 v^*(x) = \max_i x_i + \frac{1}{2} \max_{q \in [-1,1]^N} \langle D^2  v^*(x) \cdot q, q \rangle  
\end{align}
Since this PDE has not been solved explicitly for general $N$, it does not provide numerical estimates of the regret. (On the other hand, our framework provides explicit upper and lower bounds since our potentials are explicit.) 

Since $v^*$ represents a value function of a scaled game, we demonstrate its relationship to our framework by introducing scaled versions of our potentials.  The scaled lower bound potential is $\hat u^{\delta}(x)  = {\sqrt {\delta}} \hat u \left (\frac{x}{\sqrt {\delta}} \right) $ where $\hat u$ is a solution of \eqref{eq:pde_geom_lb}.  Note that $D^2 \hat u^{\delta} (x) = \frac{1}{\sqrt {\delta}} D^2 \hat u \left (\frac {x}{\sqrt {\delta}} \right)$. Thus, by \eqref{eq:pde_geom_lb_ineq},
\begin{align*}
 \hat u \left(\frac {x}{\sqrt {\delta}}\right)  \leq \max_i \frac {x_i}{\sqrt {\delta}} + \frac {1- \delta}{ 2\delta} \mathbb E_a \left \langle D^2 \hat u\left (\frac {x}{\sqrt {\delta}} \right) \cdot q, q \right \rangle 
 \end{align*} 
 which implies that 
 \[
\frac {1}{\sqrt {\delta}} \hat u^{\delta} \left(x \right)  \leq \frac {1}{\sqrt {\delta}} \max_i x_i + \frac {1- \delta}{ 2\delta} \mathbb E_a \left \langle \sqrt {\delta} D^2 \hat u^{\delta} \left (x  \right) \cdot q, q \right \rangle  
\]
Therefore, $\hat u^{\delta}$ satisfies
\[
\hat u^{\delta} \left(x \right)  \leq  \max_i x_i + \frac {1- \delta}{ 2} \mathbb E_a \left \langle D^2 \hat u^{\delta} \left (x  \right) \cdot q, q \right \rangle  
\]

Since $ \frac{1}{2} \max_{q \in [-1,1]^N} \langle D^2 \hat u^{\delta} \cdot q, q \rangle  \geq   \frac {1- \delta}{ 2} \mathbb E_a \left \langle D^2 \hat u^{\delta} \left (x  \right) \cdot q, q \right \rangle$, 
$\hat u^{\delta}$ is a so-called subsolution of the nonlinear PDE given by \eqref{eq:drenska_pde_geom}. Since these PDEs have comparison principles,  $ \hat u^{\delta} \leq  v^*$.

We also define a scaled upper bound potential  $\hat w^{\delta}$ by $\hat w^{\delta}(x)  = {\sqrt {\delta}} \hat w \left (\frac{x}{\sqrt {\delta}} \right) $ where $\hat w$ is the solution of  \eqref{eq:pde_geom_ub}. By an argument similar to the one above, $\hat w^{\delta}$ satisfies
\[
\hat w^{\delta} \left(x \right)  \geq  \max_i x_i + \frac {1- \delta}{ 2} \max_{q \in [-1,1]^N} \left \langle D^2 \hat w^{\delta} \left (x  \right) \cdot q, q \right \rangle  
\]

Since $ \lim_{\delta \rightarrow 0} \frac {1- \delta}{ 2}\max_{q \in [-1,1]^N} \left \langle D^2 \hat w^{\delta} \left (x  \right) \cdot q, q \right \rangle  =   \frac {1}{ 2}\max_{q \in [-1,1]^N} \left \langle D^2 \hat w^{\delta} \left (x  \right) \cdot q, q \right \rangle $,  in a scaling limit, $\hat w^{\delta}$ is a supersolution of \eqref{eq:drenska_pde_geom}. By the comparison principles, $v^*\leq  \lim_{\delta \rightarrow 0} \hat w^{\delta}$. 

While the above discussion provides intuition on why our framework leads to bounds, it relies on comparison principles for viscosity solutions, which are not elementary. A key advantage of the present paper is that our analysis is entirely elementary (using little more than Taylor's theorem). Our framework also is nonasymptotic,  providing bounds for any $\delta > 0$, rather than addressing only a scaling limit. 

\subsection{Relationship to Existing Bounds}
\label{sec:related_work}

\begin{figure}
  \centering
    {\includegraphics[width=0.49\linewidth]{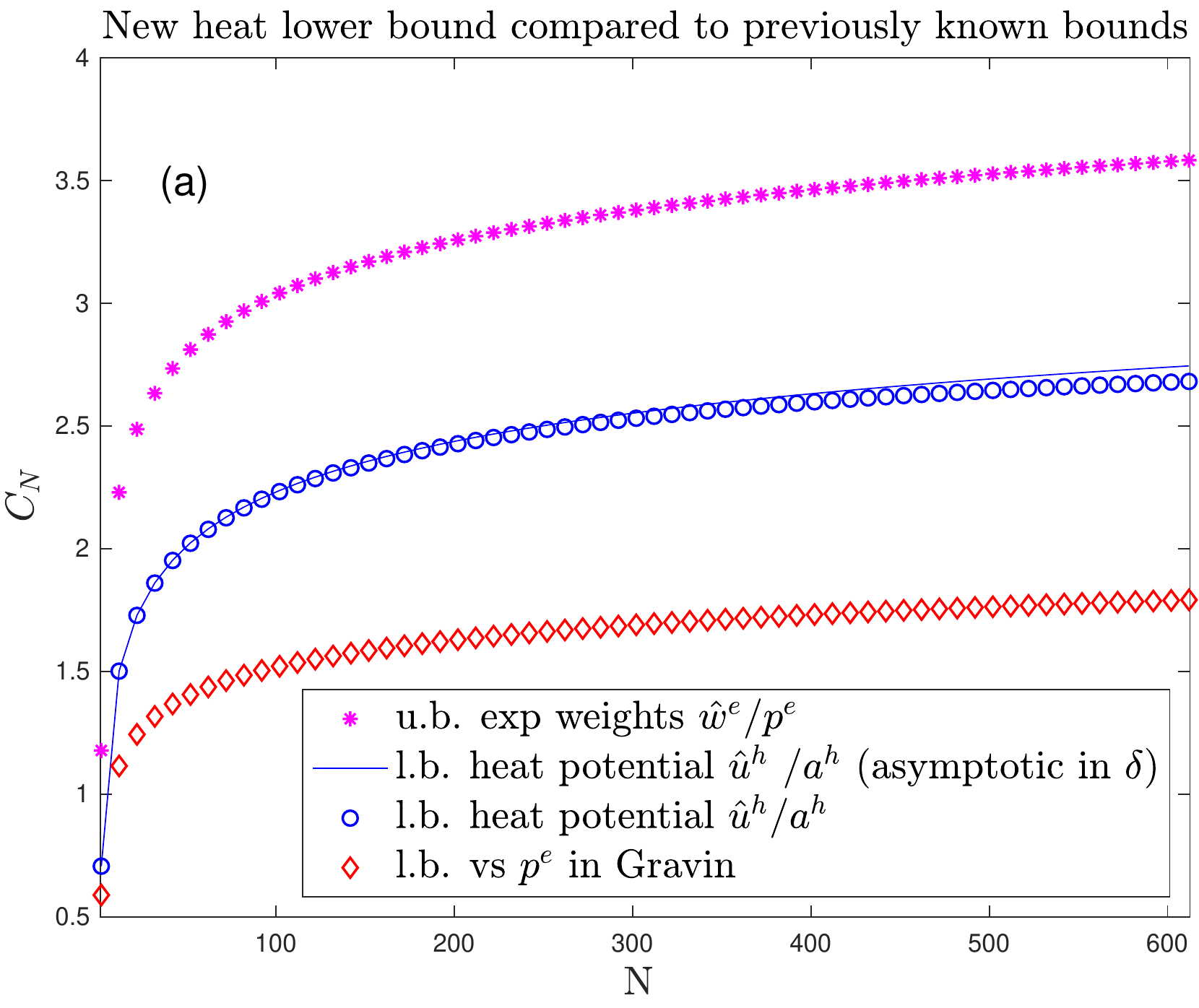}
   \phantomsubcaption\label{fig:l}}
  {\includegraphics[width=0.49\linewidth]{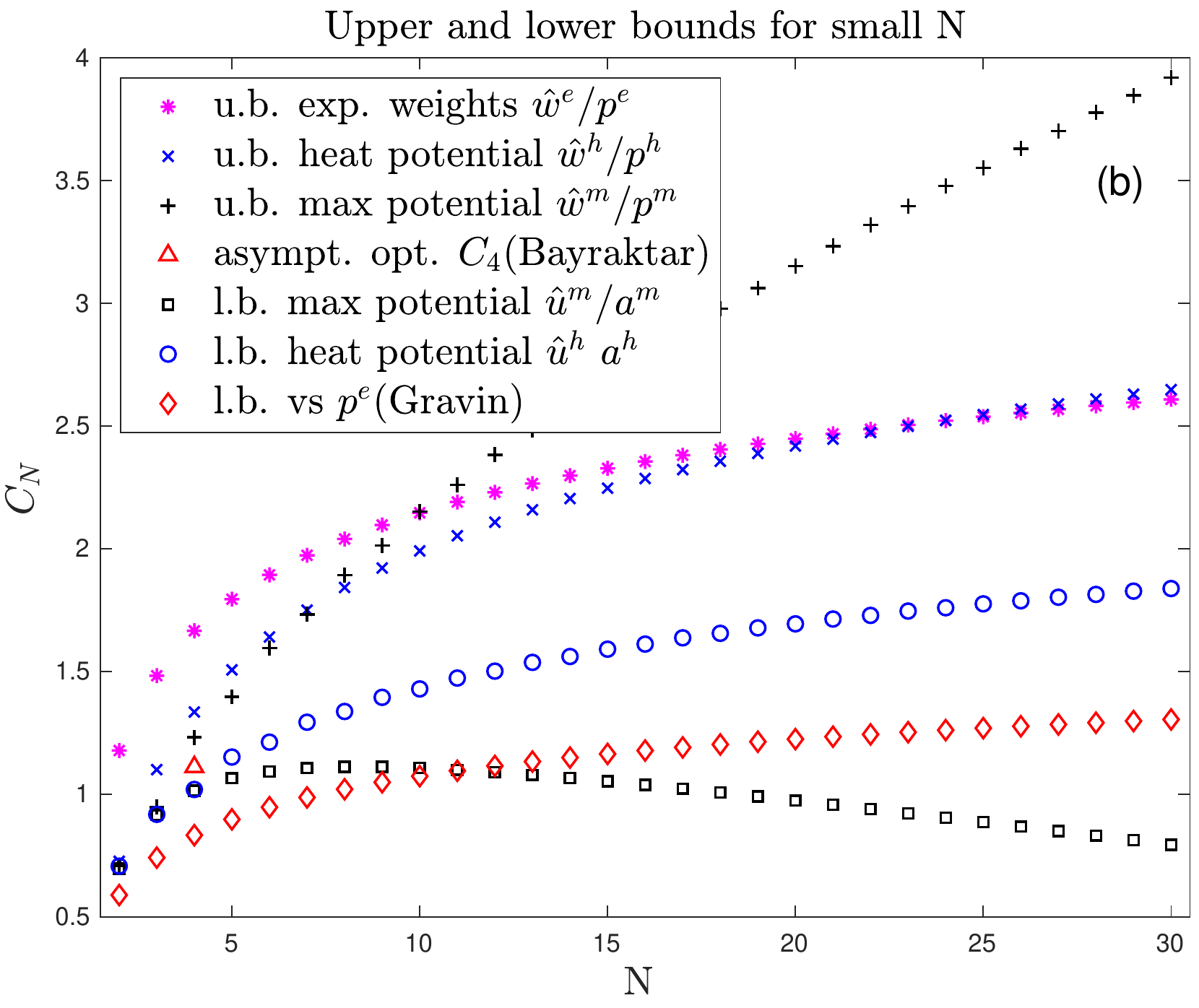}
   \phantomsubcaption\label{fig:2}}
  \caption{ Plots of $C_N$ with $N$ where $C_N = (\hat u(0) - E)\sqrt {\delta}$ for a lower bound (l.b.) potential $\hat u$ and the associated adversary $a$  (the resulting l.b. is $C_N/ \sqrt {\delta} \leq \min_p R(a,p)$) and $C_N=(\hat w(0) + E)\sqrt {\delta}$ for an upper bound (u.b.) potential $\hat w$ and the corresponding player $p$ (the resulting u.b. is  $\max_a R(a,p) \leq C_N/ \sqrt {\delta}$). Each $C_N$ is determined for $\frac{1}{\delta} = 10^6$ except where it is specified to be asymptotic. Plot \subref{fig:l} compares the new heat-based l.b. with previously known bounds and plot \subref{fig:2} shows the u.b. and l.b.'s for small $N$.}
  \label{fig:CN}
\end{figure}

While there are several known lower bounds for regret in the fixed horizon setting, our lower bounds seem to be the first such results in the geometric stopping setting with an arbitrary number of experts.  A lower bound is, however, known when the player's strategy is fixed to be exponential weights (the strategy we have called $p^e$); indeed, in this setting \citet{gravin17} show that the regret (rescaled to our setting of losses in $[-1,1]$) is asymptotically at most $\sqrt{\frac {\log N}{2 \delta}}$ as $\delta \rightarrow 0$. It is interesting to compare that result with the lower bound we proved using the heat adversary $a^h$. Since $\lim_{N \rightarrow \infty} \frac{\max_iY_i}  {\sqrt {2 \log N}} =1$ for an $N$-dimensional Gaussian random vector $Y$ with mean zero and identity covariance matrix.  
\begin{align}
\label{eq:ewap_ratio}
\lim_{N \rightarrow \infty} \frac{1}{\sqrt {2 \log N}} \lim_{\delta \rightarrow 0} \sqrt{\delta} [\hat u^h(0) - E_{a^h}] =  \frac{\sqrt{\pi}}{2} \approx .89
\end{align}
This shows that in the limit where $\delta \rightarrow 0$ first and then $N \rightarrow \infty$, the lower bound for our heat adversary $a^h$ is tighter than the lower bound of \citet{gravin17} (for which the limit analogous to \eqref{eq:ewap_ratio} is $\frac{1}{2}$).
In the nonasymptotic settting, for sufficiently small $\delta$, our lower bound also improves on the 
one of \citet{gravin17}, as shown in Figure \ref{fig:CN}\subref{fig:l}.\footnote{Note that $\mathbb E_Y \max Y_i = \int_{-\infty}^\infty t \frac {d}{dt} \Phi (t)^N dt$ where $\Phi$ is the c.d.f. of the Gaussian random variable $N(0,1)$. Therefore, for comparison purposes, we evaluate the expectation of the maximum of Gaussian using numerical integration (\textit{integral} function in MATLAB).}     (Moreover, our guarantee is given with respect to an arbitrary player strategy and not just $p^e$.)
 
Our upper bound $\sqrt{\frac{2(1-\delta)}{ \delta} {\log N}}$ for  $p^e$ also improves upon the corresponding upper bound $ \sqrt{\frac{2}{ \delta} {\log N}}$ determined in \citet{gravin17} (as rescaled for our $[-1,1]^N$ losses).  Furthermore, when $ N$  and $\delta$ are small, as illustrated by Figure \ref{fig:CN}\subref{fig:2}, the max-based player $p^m$ and the heat- based player $p^h$ improve the upper bounds attained by $p^e$  (the heat adversary $a^h$ also remains tighter than the corresponding lower bound in \citet{gravin17}).

\section{Conclusion}

In this work, we have extended the framework of \citet{kobzar} for the expert problem from the fixed horizon setting to the geometric stopping one. This framework uses potentials satisfying certain PDE-style inequalities, and it gives lower as well as upper bounds. 

Our lower bounds appear to be the first such results in the geometric setting for general $N$. The leading order behavior of the lower bound based on the Laplace transform of the solution to the linear heat equation (heat potential) is $\Omega \Big(\sqrt {\frac {\log N}{\delta}} \Big)$, which matches up to a prefactor the leading order upper bound obtained using exponential weights.  Our lower bound is associated with a simple randomized strategy for the adversary that is independent of the accumulated regret or any other history. 

Also our upper bounds based on the heat potential and a new max potential are tighter for small $N$ than those guaranteed by the exponentially weighted average forecaster.  

\acks{V.A.K and R.V.K. are supported, in part, by NSF grant DMS-1311833. V.A.K. is also supported by the Moore-Sloan Data Science Environment at New York University. }

\bibliography{expert_bounds_article}
\appendix

\section{Proof of Theorem \ref{thm:geom_lb}}
\label{app:thm_geom_lb}

We show that $\hat u$ bounds $v_a$, modulo an ``error" term, in two steps. First, we define a new problem, which is the same as the geometric stopping problem except that it starts at a given time $t \leq -1$, and ends at time $0$ (if it does not end sooner in accordance with the geometric stopping condition) and show that the value of the new problem approaches $v_a$ as $t \rightarrow -\infty$. Second, we show that $\hat u$  bounds below the value of the new problem for all $t$ (modulo an ``error" term as well).

\subsection{Convergence of the Value of the New Problem to the Original One}
Similarly to \eqref{eq:lb_functional_equation}, the value function $g$ of the new problem is characterized by the following DP:
\begin {align*}
g(x,0) & = \max_{ i} x_i \\
g(x,t) &=  \delta \max_{ i } x_i + (1-\delta) \min_{p}  \mathbb E_{a,p} ~g(x+r,t+1) ~\text{if}~ t \leq -1
\end{align*}
It has an equivalent definition given by $g(x,t) =  \min_p \mathbb E_{G, p,a} \max_i \Big (x_i+ \sum_{\tau \in [\max(t,T)]} r_{i,\tau} \Big)$.  For these purposes, the geometric random variable $G$ is the same as the standard geometric distribution except that it is multiplied by $-1$ so that its outcomes are negative numbers (consistently with our time counting convention). 

Since in each period the regret can decrease by at most 2, from the definition of $v_a$ in \eqref{eq: geom_adversary_value}, we obtain $g(x,t) -s(t) \leq v(x) $ for any $t \leq 0$
where 
\begin{align*}
s(t) = \mathbb E_G  \Big [\mathbb 1_{T<t}  \sum_{T<t} 2 \Big]  =  2 \sum_{\tau = -t+1}^{\infty} (1-\delta)^{\tau-1} \delta ( \tau +t ) =   \frac{2 (1-\delta)^{-t}} {\delta} 
\end{align*}
Since $\lim_{t \rightarrow  -\infty} s(t) =0$, it suffices to bound $g$ below.  

\subsection{Lower Bound on the Value of the New Problem}
Let the error term of the new problem be given by $E(t) =   (1- \delta)^{-t} E(0)+  \sum_{\tau \in [t]} (1- \delta)^{|\tau|}K$  for $t \leq -1$ where $K$ is any constant, satisfying $\frac {1}{6} \text{ess sup}_{y \in [x,x-q]} D^3\hat u (y)[ q,q,q]   \leq  K$ uniformly for  all $q$ in the support of $a$ and all $x $. Note that $ \lim_{t \rightarrow  -\infty} E(t) = \frac {1-\delta}{\delta}K = E$.

To bound $g$ below, we show that  $\hat u(x) -  E(t) \leq g(x,t) $ for all $x \in \mathbb R^N$ and $t \leq 0$. Since $g$ is characterized by a DP, we can use induction. The initial step $\hat u (x) - E(0) \leq \max_i x_i$  follows from the uniform upper bound on $\hat u(x) - \max_i x_i$. 

To prove the inductive step, as a preliminary result, we bound below the  difference $\min_{p} ~ \mathbb E_{a, p} ~[\hat u  ( x + r)]  - \hat u(x) $ in terms of $K$   Since $\hat u$ is $C^2$ with Lipschitz continuous second-order derivatives, we use Taylor's theorem with the integral remainder:
\begin{align}
\min_{p} ~ \mathbb E_{a, p} ~[ &\hat u  ( x + r)]  - \hat u(x) = \min_{p} \mathbb E_{a, p} ~[\hat u  ( x - q)] +q_{I}] - \hat u(x) \nonumber \\
=\min_{p} \mathbb E_{a, p}    & \Big [ q_{I} -   \langle q, \nabla \hat u ( x )  \rangle \label{eq:diff_2} 
 +   \frac {1}{2}\langle D^2 \hat u(x) q, q \rangle -   \int_0^1 D^3\hat u(x-\mu q)[q,q,q]\frac{(1-\mu)^2}{2}d\mu \Big ]   \\ 
 \geq \frac{\delta}{1- \delta} (& \hat u(x)-  \max_i x_i)-K \nonumber
\end{align}
As discussed in Section \ref{sec:intro}, $q$ distributed according to $a$ and $I$ distributed according to $p$ are independent conditioned on history. Therefore, $\mathbb E_{p,a}  [q_{I}-   \langle q, \nabla \hat u ( x )  \rangle] = \langle p - \nabla \hat u(x), \mathbb E_{a} q \rangle = 0$ for all $p$ since $a$ is balanced and  $\sum_{i} \partial_i  \hat u =1$ by linearity of $\hat u$ along $\mathbb 1$. We also used the condition on the potential \eqref {eq:pde_geom_lb_ineq}.  Rearranging \eqref{eq:diff_2}, we obtain
 \[
 (1-\delta) (\min_{p} ~ \mathbb E_{a, p} ~[\hat u  ( x + r)]  - \hat u(x)) - \delta (\hat u(x) -  \max_i x_i)+(1-\delta)K \geq 0
\]
Using this inequality, the inductive hypothesis $u(x+ r) -E(t+1) \leq g(x+ r , t+1)$, and the dynamic programming characterization of $g$, we obtain
\begin{align*}
 \hat u  (x )  - E(t) &\leq \hat u(x)  - \delta (\hat u(x)  - \max_i x_i) + (1- \delta) \min_p ~ \mathbb E_{a, p}   [\hat u(x+ r ) - \hat u(x)]  +  (1- \delta) K -E(t)\\
& \leq \delta \max_i x_i + (1- \delta) \min_p  ~ \mathbb E_{a, p}   [\hat u(x+ r )]  +  (1- \delta) K  - E(t) \\
& \leq \delta \max_i x_i + (1- \delta) \min_p  ~ \mathbb E_{ a, p}  [g(x+ r ,t+1) + E(t+1)] +  (1- \delta) K  - E(t) \\
&=g ( x,  t) 
\end{align*}
Finally, observe that $E(t)$ satisfies the recursion $E(t) = (1- \delta) (E(t+1) +K )$ used in the last equality. 

\section{Proof of Proposition \ref{rem:geom_lb_lipschitz} }
\label{app:geom_lb_lipschitz}
If $D^3 \hat u$ exists and is Lipschitz continuous, then \eqref{eq:diff_2} can be replaced by
\begin{align*}
\min_{p} \mathbb E_{a, p}  \Big [&q_I -\nabla \hat u(x)\cdot q+\frac{1}{2}\langle D^2 \hat u(x)\cdot q,q\rangle \\
&-\frac{1}{6} D^3 \hat u(x) [q,q,q]+\int_0^1 D^4\hat u(x-\mu q)[q,q,q,q]\frac{(1-\mu)^3}{6}d\mu \Big]
\end{align*}
Since the adversary $a$ is symmetric, $q$ has the same distribution as $-q $. Therefore,  $\mathbb E_{a}q_iq_jq_k =0$ for any $i$, $j$, and $k$ and consequently $\mathbb E_{a}D^3\hat u(x)[q,q,q]=0$.  The remainder of the proof of Theorem \ref{thm:geom_lb} is the same except that we  define $K$ as any constant  satisfying $ -  \frac {1}{24}\text{ess inf}_{y \in [x,x-q]} ~D^{4}\hat u(y) [q,q,q,q]   \leq  K$ uniformly for all $q$ in the support of $a$ and all $x$.

 \section{Proof of Theorem \ref{thm:geom_ub}}
\label{app:thm_geom_ub} Since $p=\nabla \hat w(x)$ and $I$ and $q$ are distributed independently,  $\mathbb E_{a,p}   [  q_I  -\nabla \hat w(x)\cdot q] = \mathbb E_{a}   [ \langle p, q \rangle -\nabla \hat w(x)\cdot q] = 0 $ for all $a$, which also eliminates the first-order derivative from the Taylor expansion. Therefore,
\begin{align}
&\max_{a} ~ \mathbb E_{a, p} ~[ \hat w  ( x + r)]  - \hat w(x) \nonumber \\
&=\max_{a} \mathbb E_{a}     \Big [  \frac {1}{2}\langle D^2 \hat w(x) q, q \rangle -   \int_0^1 D^3\hat w(x-\mu q)[q,q,q]\frac{(1-\mu)^2}{2}d\mu \Big ]  \label {eq:geom_ub_evolution}\\ 
& \leq \frac{\delta}{1- \delta} ( \hat w(x)-  \max_i x_i)+K \nonumber    
\end{align}

The rest of the proof of Theorem \ref{thm:geom_ub} is similar to the proof of Theorem \ref{thm:geom_lb}.

\section {Proof of Proposition \ref{rem:fixed_to_geom_ub_const_w_t} }
 \label{app:fixed_to_geom_ub_const_w_t} 
 
In this setting,  the geometric upper bound potential $\hat w$ is given by 
\begin{align*}
\hat w(x) =  \int_{- \infty }^{0} e^{t} w(x,t) dt + C =  \Phi (x) - k  + C 
\end{align*}
which implies that  $\langle D^2 \hat w  \cdot q, q \rangle= \langle D^2 w  \cdot q, q \rangle  = \Phi (x)$. Note $C$ is a constant that such  $\max_i x_i - w(x, 0)   \leq  C $ for all $x$.  

This fact, the fact that  $w(x, 0) = \Phi (x)$, and the properties of the fixed horizon upper bound potential $w$, in particular \eqref{eq:pde_ineq_fixed_ub} and \eqref{eq:pde_final_fixed_ub}, imply that for two arbitrary points $x$ and $y$ in $\mathbb R^N$ 
\begin{align*}
 \frac{1-\delta}{2\delta} \langle D^2 \hat w(y) q, q \rangle -  \hat w(x)+  \max_i x_i =
\Big (\frac{1-\delta}{2\delta} \langle D^2  w(y) q, q \rangle + k \Big) + \Big ( \max_i x_i  -\Phi (x) \Big)  - C  
 \leq 0
\end{align*}
or equivalently $ \frac{1}{2} \langle D^2 \hat w(y) q, q \rangle  \leq   \frac{\delta}{1-\delta} (\hat w(x)-  \max_i x_i)$. Therefore, in the proof of  Theorem \ref{thm:geom_ub} in Appendix \ref{app:thm_geom_ub}, we can use  Taylor's theorem with the mean value form of the second-order remainder. Accordingly, \eqref{eq:geom_ub_evolution}  is replaced by 
  \begin{align*}
&\max_{a} ~ \mathbb E_{a, p} ~[ \hat w  ( x + r)]  - \hat w(x) \nonumber =\max_{a} \mathbb E_{a}  \frac {1}{2}\langle D^2 \hat w(y) q, q \rangle   \leq \frac{\delta}{1- \delta} ( \hat w(x)-  \max_i x_i)   
\end{align*}
Note that the error term $K$ has been eliminated from this expression. 

\section{Heat Potential: Final Time Bound}
\label {heat_final_period}

Since $ - \max_i (x-y)_i \geq -\max_i x_i + \min_i y_i$, we have  
\begin{align*}
\varphi(x,0) - \varphi(x,-\delta)   &=    \alpha \int e^{- \frac{\|y\|^2}{2\sigma^2}}  \max_i x_i - \max_i (x_i-y_i)   dy \\
&\geq \alpha\int e^{- \frac{\|y\|^2}{2\sigma^2}}  \min_i y_i dy  = -\sigma \mathbb E \max_{ i}  Y_i   
\end{align*}
where  $\sigma = \sqrt {-2\kappa \delta}$. Similarly, since $ - \max_i (x-y)_i \leq -\max_i x_i + \max y_i$, we obtain $\varphi(x,0) - \varphi(x,-\delta) \leq   \sqrt  {-2\kappa  \delta} \mathbb E \max_ i  Y_i$.

\section{Max Potential: Final Time Bound}
\label {max_final_period}
We have for any $x$
\[
\psi(x,0) - \psi(x,-\delta)  = x_{(1)} -\frac {1}{N} \sum_{l=1}^N x_{(l)} -    \sqrt {2 \kappa \delta} \sum_{l=1}^{N-1} c_l f(z_l)
=  \sqrt {2 \kappa \delta} \sum_{l=1}^{N-1} c_l z_l -    \sqrt {2 \kappa \delta} \sum_{l=1}^{N-1} c_l f(z_l)
\]
Since $-\sqrt{\frac{2}{\pi}} \leq z -f(z) \leq 0$ for $z \geq 0$, we have $ -2\sqrt {\frac{\kappa \delta }{\pi}}  \frac {N-1}{N}  \leq \psi(x,0) - \psi(x,-\delta) \leq 0$.

\end{document}